\documentclass[letterpaper, 10 pt, conference]{ieeeconf}
\usepackage[bindingoffset=0.2in,%
            left=.75in,right=.75in,top=.75in,bottom=.75in,%
            footskip=.75in]{geometry}
\usepackage{graphicx}
\usepackage{amsmath}
\usepackage{amssymb}
\usepackage{subcaption}
\usepackage{caption}
\usepackage[dvipsnames]{xcolor}
\IEEEoverridecommandlockouts                              
                                                          
\allowdisplaybreaks
\makeatletter
\newcommand{\pushright}[1]{\ifmeasuring@#1\else\omit\hfill$\displaystyle#1$\fi\ignorespaces}
\newcommand{\pushleft}[1]{\ifmeasuring@#1\else\omit$\displaystyle#1$\hfill\fi\ignorespaces}
\makeatother

\begin{document}

\title{{\LARGE \textbf{A Hierarchical Framework for Long-term and Robust
Deployment of Field Ground Robots in Large-Scale Farming }}}
\author{Stuart Eiffert, Nathan D. Wallace, He Kong, Navid Pirmarzdashti, and
Salah Sukkarieh\thanks{%
The authors are with the Australian Centre for Field Robotics, The
University of Sydney, NSW 2006, Australia. Stuart Eiffert and Nathan Wallace contributed equally to this work. Corresponding author: \texttt%
{\small h.kong@acfr.usyd.edu.au}}}
\maketitle

\begin{abstract}
Achieving long term autonomy of robots operating in dynamic environments such as farms remains a significant challenge. Arguably, the most demanding factors to achieve this are the on-board resource constraints such as energy, planning in the presence of moving individuals such as livestock and people, and handling unknown and undulating terrain. These considerations require a robot to be adaptive in its immediate actions in order to successfully achieve long-term, resource-efficient and robust autonomy. To achieve this, we propose a hierarchical framework that integrates a local dynamic path planner with a longer term objective based planner and advanced motion control methods, whilst taking into consideration the dynamic responses of moving individuals within the environment. The framework is motivated by and synthesizes our recent work on energy aware mission planning, path planning in dynamic environments, and receding horizon motion control. In this paper we detail the proposed framework and outline its integration on a robotic platform. We evaluate the strategy in extensive simulated trials, traversing between objective waypoints to complete tasks such as soil sampling, weeding and recharging across a dynamic environment, demonstrating its capability to robustly adapt long term mission plans in the presence of moving individuals and obstacles for real world applications such as large scale farming.

\end{abstract}

\thispagestyle{empty} \pagestyle{empty}



\section{INTRODUCTION}
This work is primarily motivated by recent developments in the automation of field operations for unstructured environments such as construction, mining, and agriculture \cite{Bechar2017}, where the application of field robotics has the potential to automate many essential or desirable tasks that are often labour-intensive, tedious or dangerous. To achieve this, however, a number of significant challenges must be addressed, including robust motion control across a variety of terrains \cite{Wallace_ICRA}--\cite{Wallace_FSR}, long-term autonomy and efficiency under resource constraints \cite{Wallace_WROCO}--\cite{Wallace_Agricontrol}, and safe operation around moving obstacles \cite{Eiffert_ICRA}--\cite{Eiffert_ACRA}. In this paper we propose a principled operational framework capable of handling these challenges, building upon our recent research and development activities in agricultural robotics. That said, our proposed methodology are also applicable for other domains such as mining, infrastructure, and construction.

 \begin{figure}
    \centering
	\includegraphics[width=8.0cm,height=5.5cm]{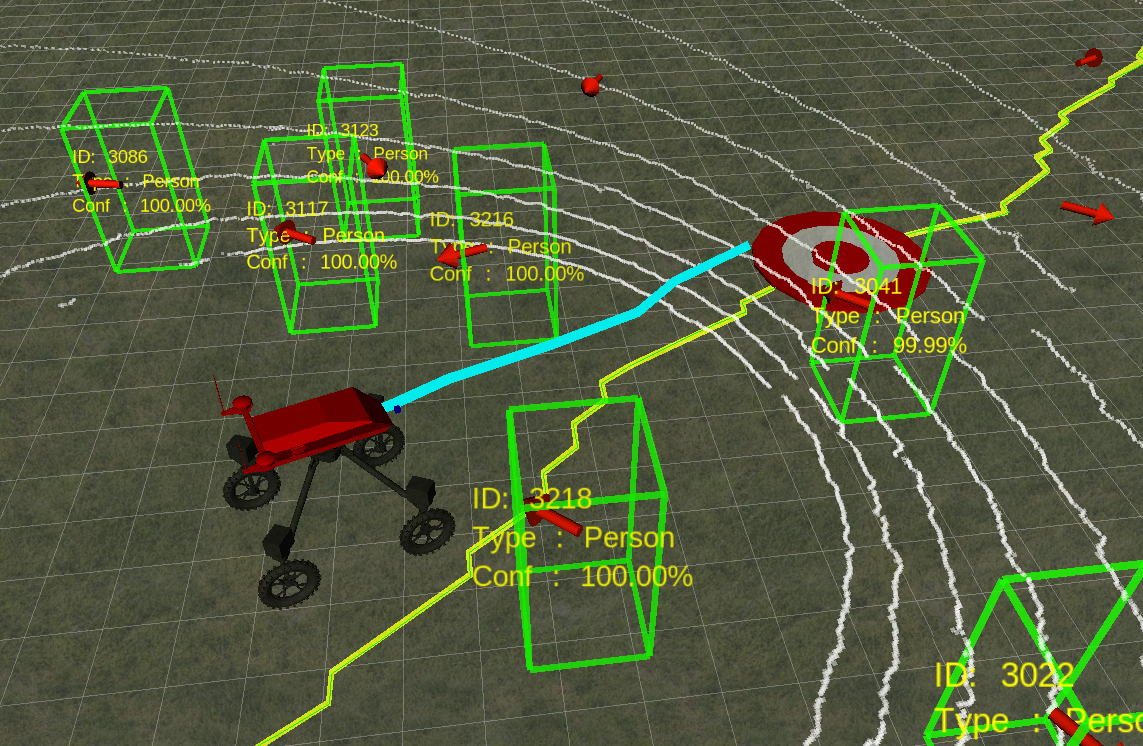}
	\caption{\textit{The hierarchical framework deployed in simulation, demonstrating the output of the local planner (blue) being used to update the long term plan (yellow) in order to navigate between mission waypoints in a dynamic environment.}}
	\label{system_overview}
\end{figure}




In agriculture, there are a large variety of essential tasks---including soil sampling, weeding, crop observation, and recharging---that are often dispersed widely over large geographical areas. In order to complete these tasks a robot can be required to navigate off-road environments in the presence of dynamic agents such as livestock and humans.
This requires the planning of longer term paths between a series of mission objective waypoints whilst ensuring the adaptability of these paths to changes in the environment. To allow the robot to efficiently travel between waypoints, any short term plan updates must be made in a way that understands how the dynamic agents are likely to move around the robot and respond to its motion. Whilst previous work has demonstrated how static and dynamic elements of an unknown environment can be learnt and mapped through explicit identification of the agents, a complete long term planner which accounts for how these agents will respond to a robot's motion as it moves through the static world has not yet been demonstrated.

In this work we detail a hierarchical framework integrating a dynamic perception pipeline, which learns both static and dynamic elements of an unknown environment, with a multi-level path planner and tracker. This path planner consists of a local dynamic path planner which considers the responses of nearby agents, and a longer-term, objective and energy-aware global planner, as well as an advanced receding horizon motion controller. We demonstrate this integrated framework in extensive simulated trials\footnote{Due to the recent breakout of COVID-19 in Australia, our planned field validation tests were postponed and will be pursued in due course.} to verify its use for long term and robust autonomy in large scale farming. These trials are conducted in an environment which replicates terrain from real world field trial locations, and are deployed on a dynamic simulation of the University of Sydney's Swagbot agricultural robot platform. Our contributions include (i) a hierarchical framework for the deployment of field robots in unstructured and dynamic environments, integrating our prior work in \cite{Wallace_ICRA}--\cite{Eiffert_ACRA}; (ii) improvement of collision avoidance around dynamic agents, extending on our recent work \cite{Eiffert_ICRA} for better planning persistence; (iii) comprehensive and high-fidelity validation in simulation, including traversal between waypoints to accomplish an updated set of tasks, whilst ensuring safe and efficient planning around dynamic agents, and adhering to resource and recharging constraints; (iv) demonstration of how this framework can be used to adapt to changing resource constraints by using different local planner approaches to achieve varied resource use.

\section{RELATED WORK}

While mobile robots have been applied to dynamic environments for decades \cite{Fox1997}--\cite{VanDenBerg2011} and see continuous improvement with regards to their ability to plan paths around moving agents \cite{Eiffert_ICRA}, \cite{Everett2018}--\cite{Tomizuka2020}, the problem of extending these systems for long term deployment, with the ability to update their mission objectives in real time, remains an outstanding challenge. Existing solutions in agriculture tend to focus on more structured problems such as completing a single task in row crops or orchards \cite{Bechar2017} and do not take into consideration the difference between static and dynamic elements in their environments, limiting their use around livestock and people. Additionally, limited work has focused on long term applications that require consideration of resource management in path planning.

\subsection{Resource Aware Path Planning}

To best utilise mobile robotic agents---particularly electric-powered wheeled mobile robots (WMRs)---in off-road and large scale environments, it is necessary to be aware of the levels of onboard resources, and the anticipated costs of performing tasks and actions in the environment. This is especially relevant for energy usage, which determines the robot's range and max operational time and has been the topic of our previous work on modelling the energy cost of WMR motion \cite{Wallace_WROCO}.

The development of energy-aware and efficient path planning methods has also received significant interest in recent years, utilising cost models for fuel and energy use for point-to-point and coverage path planning \cite{Qian2010}--\cite{Isler2018} and planning energy efficient multi-stage paths for WMRs in undulating off-road environments \cite{Wallace_Agricontrol}. The problem of achieving longer term autonomy under resource constraints is often modelled as variants of the Orienteering Problem (OP) across numerous domains \cite{Yu2016}--\cite{Faigl2017} with recent literature exploring approaches which allow for recharging of resources \cite{Laporte2013}--\cite{Cassandras2018}. Our recent work on the OP with replenishment (OPR) \cite{Wallace_CASE2020} presents a formulation which can handle multiple revisits to a number of recharging stations distributed throughout the operating environment, while also optimising the amount of time spent recharging to ensure tasks are completed as efficiently as possible---driven by the consideration that it is desirable for agricultural tasks to be completed in time.

\subsection{Dynamic Path Planning} 
The ability to operate safely around moving agents is critical for any robotic system intended for use around humans or livestock. There exists significant interest in motion prediction methods in unstructured environments for dynamic path planning \cite{Everett2018}--\cite{Xing2020}, \cite{Manocha2015}, and methods that consider the social responses in crowds and herds \cite{Eiffert_ICRA}--\cite{Eiffert_ACRA}. However, unlike applications in more structured environments such as autonomous driving \cite{Tomizuka2020}--\cite{Xing2020}, where we see local maneuver-based path planning used to update higher level mission planners \cite{Katrakazas2015}--\cite{Schwarting2018}---these works have not yet been applied to systems that require longer term autonomy, such as those addressed here.

Existing dynamic path planning methods for long term field robot autonomy are limited to simple collision avoidance systems, such as velocity obstacles \cite{Fiorini1998} and potential fields \cite{Ge2002}, which we compare against as an alternative to the local path planning module developed here. Also, these methods are generally applied to semi-structured environments, such as indoor use, rather than large scale farming, where current implementations make use of fail safe (FS) methods which simply stop the robot, or control its speed along a predefined path in the presence of moving individuals. Whilst these methods allow certain operations around moving agents, they tend to greatly reduce the efficiency of task completion and can result in the robot becoming stuck with no feasible path forward.
 \begin{figure*}
    \centering
	\includegraphics[width=16.5cm,height=4.5cm]{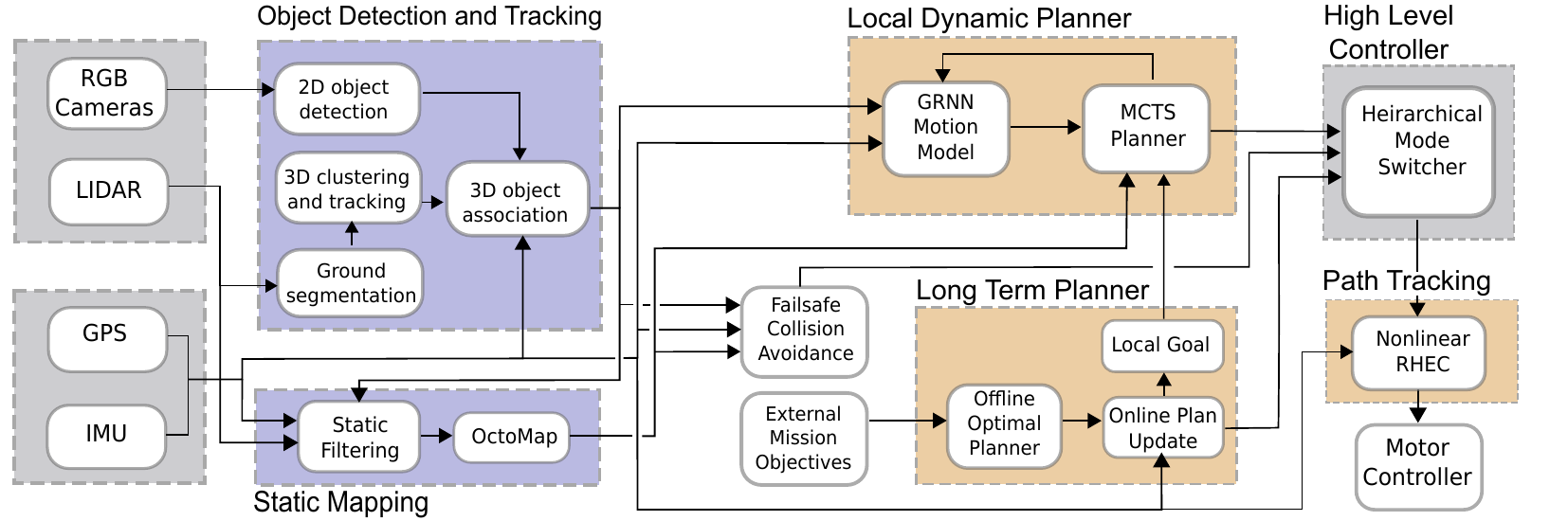}
	\caption{\textit{System overview of the hierarchical framework, illustrating the communication between each planner, the failsafe collision module, and the high level controller. Mission objectives are provided externally to the long term planner.}}
	\label{system_overview}
\end{figure*}

\section{THE PROPOSED HIERARCHICAL FRAMEWORK}
Our approach combines a long term mission planner with a local dynamic path planner to achieve long term autonomy of a robot travelling between updated objective waypoints in the presence of both static obstacles and dynamic agents. This framework: (1) utilises our prior work in energy efficient path planning \cite{Wallace_WROCO}--\cite{Wallace_Agricontrol} for generating long term plans; (2) integrates this with our prior work on using Generative Recurrent Neural Networks (GRNN) and Monte Carlo Tree Search (MCTS) \cite{Eiffert_ICRA}--\cite{Eiffert_ACRA} as a local path planner; and (3) adopts online, slip-compensating Receding Horizon Motion Control (RHMC) \cite{Wallace_ICRA}--\cite{Wallace_FSR} path tracking. Additionally, our framework generates a  real-time updated map of static obstacles and traversable terrain in the environment, used both by the local dynamic planner and by a FS collision avoidance module. Fig. \ref{system_overview} outlines how each component integrates into the hierarchical framework, with the output of the local, long term and fail safe planners being passed to a high level controller.


\subsection{Long Term Mission Planner}
\label{section:missionplanner}

The long term mission plan is generated from a set of periodically updating, externally provided mission objective waypoints which must be visited for the completion of tasks such as weeding, soil sampling and recharging of the robot. These target locations are then connected into a Probabilistic Roadmap (PRM) which describes a set of kinematically feasible paths for the robot over the environment. The minimum energy paths between all pairwise combinations of locations---calculated using the energy cost of motion model developed in \cite{Wallace_WROCO}---are used to generate a `goal graph', over which an asymmetric travelling salesman problem is solved to yield the optimal ordering of visits and the corresponding energy-minimising global plan. The approach in this paper builds upon our work in \cite{Wallace_Agricontrol} by constraining the robot to operate in Ackermann configuration, thereby forbidding motion in any directions without adequate perception coverage for the sake of safety. Each sample pose in the PRM was therefore connected by Clothoid curve segments, to generate a continuous curvature path that is easily trackable by such a vehicle.

\subsection{Object Detection and Static Mapping}  
\label{section:detectionmapping}
\smallskip 
The differentiation of traversable terrain, static obstacles and dynamic agents is another essential element required for long term autonomy in unstructured environments. Whilst this problem has been addressed in more structured applications such as on road autonomous vehicles, it remains a significant challenge in unknown and unstructured terrain such as is often encountered on farms. In this work we apply a multimodal perception system that combines 3D LIDAR and 2D RGB vision to both detect dynamic agents and learn a static map of the robot's environment. This pipeline involves the identification of ground sets within the pointcloud and clustering of non-ground sets, based on work from \cite{Underwood2011}, and then association of these clustered 3D sets to 2D detected bounding boxes within the camera frame. This is achieved through the use of a Single Shot Multi-Box Detector (SSD) CNN \cite{Liu2016} and accurate calibration of the 2D and 3D sensors \cite{Nebot2019}, which allows projection of the 3D sets into the 2D frame for nearest neighbour association with the 2D bounding boxes. Intersection over union is also computed between all 2D bounding boxes and the minimal fit bounding box for projected 3D sets, which is used to validate the association and provide a classwise probability score. The remaining non-associated 3D sets are passed to a probabilistic OctoMap framework \cite{Hornung2013} for the continuous updating of a map of static obstacles and traversable terrain. This map is used for fail safe collision avoidance and as an input to our MCTS dynamic path planner, constraining the search space.

\subsection{Local Dynamic Planner}
\label{section:localplanner}

The local path planning module used in this work is adapted from our prior work using GRNNs and MCTS for planning in dynamic environments \cite{Eiffert_ICRA}. This planner uses a learnt model of social response to predict crowd dynamics during planning across the action space and has been modified in this paper for improved persistence of paths between planning iterations, and to better account for collisions between the robot and dynamic agents in continuous time between discrete planning timesteps. This section details the specific implementation of our GRNN and MCTS local path planner, however the hierarchical framework has been designed to be agnostic with regards to the specific local path planner used, as demonstrated by our comparison of various local path planners in Section~\ref{section:experiments}.

\subsubsection{Predictive Model}
Our local planner uses a learnt model of social response, predicting the future motion of each agent based on an observation of its past motion and the known robot's motion from the same time period, as well as using the planned future action of the robot as input. This model is based on our prior work \cite{Eiffert_ICRA}, and similarly uses the robot's position at the next timestep to represent its current intended action. The predictive model consists of a recurrent Encoder-Decoder framework using long short-term memory (LSTM) layers, where the output of the Encoder becomes the state of the root node used by the MCTS planner, as detailed in Section~\ref{section:mctsplanner}. The Decoder is then used by the MCTS at each simulation step as a state transition model.
Training of this model is performed as per \cite{Eiffert_ICRA}, using generated trajectories of interacting agents modelled using the  Optimal Reciprocal
Collision Avoidance (ORCA) motion model \cite{VanDenBerg2011}. 

\subsubsection{Sampling Based Planner}
\label{section:mctsplanner}
An adapted MCTS is used to search the robot's future action space to find the sequence of actions that minimises the cost shown in Eq. \ref{eqn:costfn}, where $R^t$ refers to the robot's position at time \textit{t}, $G$ is the local goal position, $N$ is the total number of agents currently observed by the robot and $U$ refers to the uncertainty of agent $i$'s estimated position at the future timestep $t$ within the search. This value is determined from the 2D Gaussian prediction for the agent's position at $t$, as detailed in \cite{Eiffert_ICRA}. $\alpha$ is a scaling parameter based on the distance between $R^t$ and the agent's position $X_i^t$, as shown in Eq. \ref{eqn:alpha}. A value of 0 is used for $\alpha$ when the agent is beyond a distance threshold of $d$, which is set to 3m for all experiments. 

This approach extends the Parallel Single Step Simulation MCTS detailed in Alg. 1 of \cite{Eiffert_ICRA} for improved persistence of plans between timesteps. Before each planning step, we check if the computed tree from the prior step can be reused to seed the current search. A comparison is made between the current observed state, and the child node from the last tree's root node which best matches the actual action taken by the robot over the last timestep.

If the positions of all nearby agents within a distance threshold of $2d$ of the robot's position in the root node are within an error $\epsilon$, equal to twice the expected sensor noise standard deviation (0.25m), we reuse the values of the node's tree to seed our current root. The reused values are scaled by a factor of $\gamma$, determined by Eq. \ref{eqn:scalefactor}, where M is the number of agents within the $2d$ distance threshold, and $P_i^t$ is the current predicted position of agent $i$. 

\begin{flalign}
Cost &= (R^t - G)^2 + \sum_{i}^{N} \alpha {U_i^t}\label{eqn:costfn}\\
\alpha&= 
\begin{cases}
    \frac{1}{X_i^t - R^t} ,& \text{if } X_i^t - R^t\leq d\\
    0,              & \text{otherwise}
\end{cases}\label{eqn:alpha} \\
\gamma &= ( \sum_{i}^{M}\frac{\epsilon-|X_i^t-P_i^t|}{\epsilon} ) / {2M}\label{eqn:scalefactor}
\end{flalign}  

Additionally, we extend the MCTS planner to better identify invalid actions with regards to collisions with both the static map, and dynamic agents between discrete timesteps.  
Similarly to \cite{Eiffert_ICRA}, our MCTS planner uses a static map, detailed in Section~\ref{section:detectionmapping}, to constrain the action space of the robot, which has been dilated by the radius of the robot. Valid actions are determined as those that do not cause a collision between the robot and the dilated map, or the robot and the position of each agent dilated by the sum of the robot's radius and the average agent radius. We extend this by comparing the straight line path connecting the robot position in parent and child nodes of two subsequent timesteps, ensuring that this line does not intersect either the contour of a static obstacle, or any other line connecting the predicted positions of dynamic agents in the same two timesteps.


\subsection{High Level Control and RHMC}
\label{section:switcher}

As in our prior work \cite{Wallace_ICRA}--\cite{Wallace_FSR}, a receding horizon estimator (RHE) is adopted to provide an estimate of the robot state and slip conditions of the terrain. Then, based on the proximity to detected dynamic agents and static obstacles, a hierarchical mode switching module---a crucial element in the integration of the global optimal planner and the local dynamic planner---determines whether to source the local reference trajectory from the dynamic planner, or to follow the online update of the global path provided by the long term planner. In the absence of obstacles, the default path tracking behaviour will compute a reference based on the robot's progress along the global path and the specified nominal speed. If a dynamic agent or static obstacle is detected within the planning area, the local dynamic planner will be engaged.

The chosen path is then passed to the RHMC module, which solves the nonlinear optimisation problem to determine the set of control actions necessary to track the desired path over the forward horizon. This module also compensates for the impact of varying slip conditions in the underlying terrain which may otherwise adversely impact the robot's motion, and potentially risk a collision with an obstacle or dynamic agent.


As the local planner operates with a planning timestep of approximately 300ms we also make use of a FS collision avoidance module, ensuring that the robot is able to reflexively react to rapid changes in its environment without having to wait for the local planner to complete planning.

\section{EXPERIMENTS}
\label{section:experiments}
This section presents extensive and high-fidelity simulation studies to validate the performance of our proposed framework, conducted over a large set of generated problem scenarios. Each scenario involves the robot departing from and returning to a recharging station, while being externally provided with a set of mission objective waypoints. Between 5--12 waypoints are randomly selected with an average spacing of 25m. The robot is also provided with an elevation map of the terrain, provided externally from aerial LIDAR data. The robot must initially plan a long term path offline that visits every waypoint, whilst ensuring that the energy constraints of the robot are adhered to by returning to the recharging station as required.

In each of these generated scenarios, we have compared our framework as shown in Fig. \ref{system_overview} to a version in which the local planner has been replaced with either a potential field (PF) based approach, as described in \cite{Eiffert_ICRA}, or with no planner, instead relying only on the FS Collision Avoidance module. We simulate dynamic agents in each scenario, interacting with the robot as it navigates to various waypoints, and have repeated all scenarios with changes to parameters such as agent density and required positional accuracy at waypoints, as described in the sequel.

Furthermore, as outlined in Section \ref{section:missionplanner}, the robot has been kinematically constrained to operate in Ackermann configuration for the purpose of ensuring that motion is only possible within the forward facing planning area dictated by the field of view of the robot's sensors. This is necessary to ensure that Swagbot---a platform with omnidirectional motion capabilities---does not move in any directions not covered by the sensor footprint, where it would otherwise risk damaging or injuring property, livestock or personnel.

 \begin{figure}
    \centering
	\includegraphics[width=8.5cm,height=5.5cm]{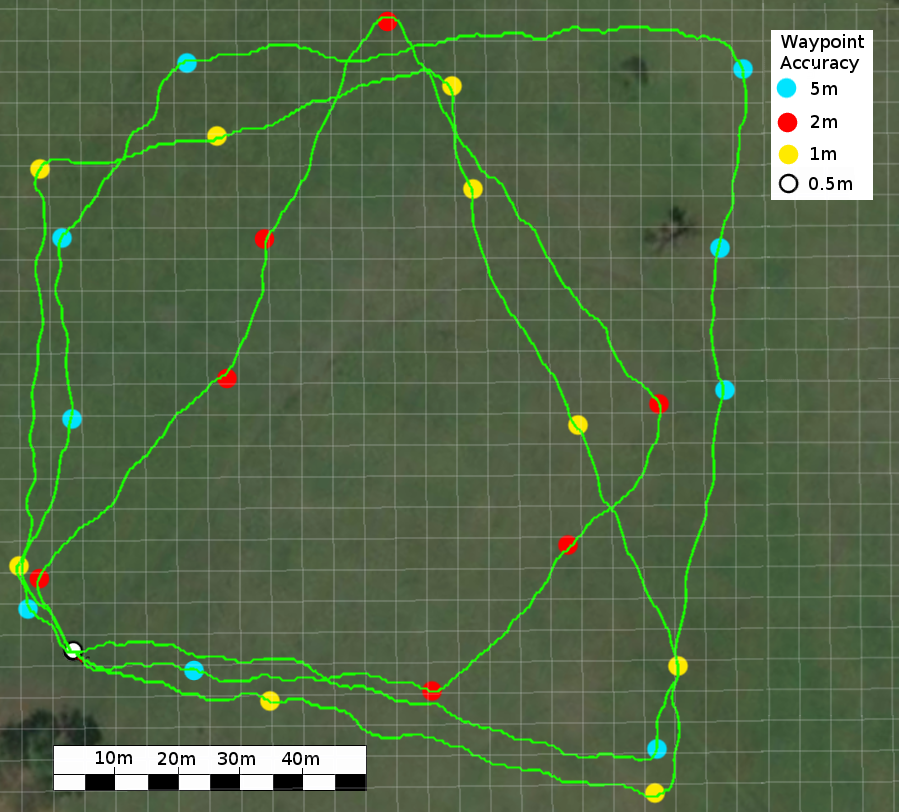}
	\caption{\textit{Example generated scenario showing offline computed paths for 3 iterations, each using a different type of waypoint with required positional accuracies of 5, 2 and 1m. Each plan begins and ends at the recharging station, with required accuracy of 0.5m, to adhere to the energy budget. These plans are computed for real world terrain generated from the Sydney University farm at Bringelly.}}
	\label{scenario_image}
\end{figure}

\subsection{Benchmark Scenarios}
We evaluate our approach using varied types of mission waypoints, each with different positional accuracy requirements for the robot. These include:
\begin{itemize}
    \item Observation Waypoint: 5m Accuracy
    \item Soil Sampling Waypoint: 2m Accuracy
    \item Weed Spraying Waypoint: 1m accuracy
    \item Recharging Station: 0.5m accuracy
\end{itemize}

Each generated scenario requires visiting a number of waypoints of a single type, whilst ensuring that the robot returns to a recharging station within its energy budget. Upon recharging, an updated set of waypoints is provided to the robot---consisting of either observing, sampling, or weeding tasks---and the robot begins again.
This iteration is repeated 8 times for each scenario. We measure performance using the following metrics:
\begin{enumerate}
    \item Deviation from the optimal path;
    \item Average speed to reach each waypoint;
    \item Number of near-collisions with dynamic agents.
\end{enumerate}

The optimal path connecting each node along the robot's longer term plan is determined by the offline planner as described in Section \ref{section:missionplanner}.
The average speed to reach a waypoint is determined by distance of the waypoint at the time that the robot receives it as its next goal, over the total duration required for the robot to achieve the required positional accuracy for that type of waypoint.

 \begin{figure*}
    \centering
	\includegraphics[width=17.5cm,height=7cm]{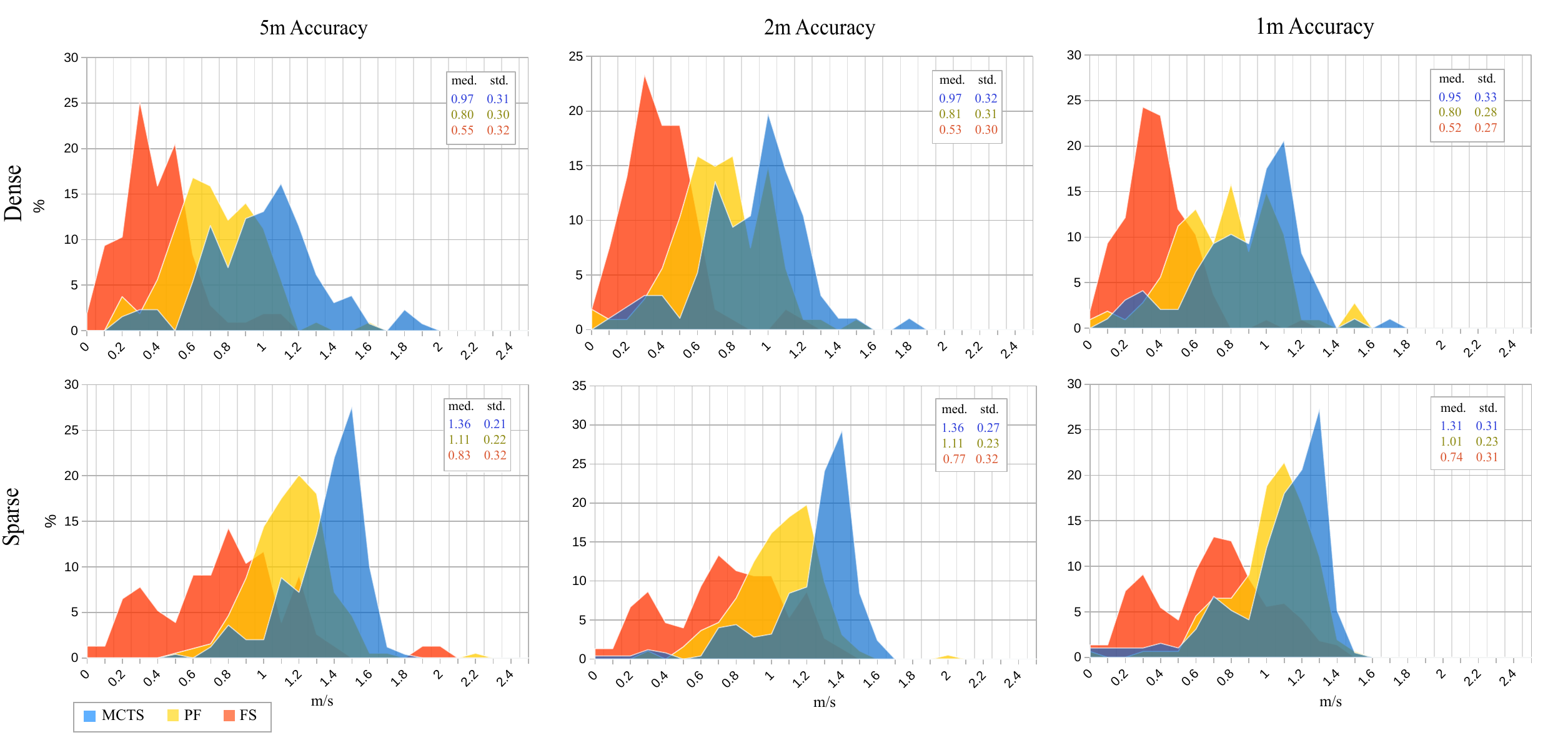}
	\caption{\textit{Average speed to achieve a required positional accuracy of 5, 2 and 1m for \% of waypoints tested. In both dense (top) and sparse (bottom) environments, the hierarchical framework is able to achieve the required accuracy in a shorter time using the more responsive local path planners, MCTS and PF, compared to the FS method.}}
	\label{avgspeed_vs_posacc}
\end{figure*}

\subsection{Implementation}

These agents are simulated using the ORCA \cite{VanDenBerg2011} model. They are spawned at random locations between waypoints for each scenario and are assigned similarly random goal points to travel towards at speeds ranging from 0.1--1.5 m/s. For all scenarios, agents are simulated with a radius of 0.5m and react to the robot assuming a radius of 1.5m.

The robot's perception has been simulated to reflect the localisation and object detection of the real world agricultural robot Swagbot, described in \cite{Eiffert_ICRA}. Sensor noise has been applied to the simulated GPS and IMU sensors used for localisation, and to the simulated obstacle detection proportional to the distance from the robot, matching that observed in previous real world experiments. Simulated terrain has been included, using information from publicly available satellite data. For the purposes of these simulated experiments we have also assumed the availability of a static obstacle map for use in the local planner, rather than the creation of one from sensor input.

We repeat each scenario for the compared methods, using either our MCTS approach, a PF planner, or only the FS method. Additionally, we repeat these experiments using two different crowd densities of 10$m^2$ (dense) and 50$m^2$ (sparse) per agent, giving on average distances of approximately 3m and 7m between each agent respectively. Simulations are run in real time, taking on average 2.5 hours to complete all iterations per scenario, depending on the type of local planner used. We additionally generate 3 separate scenarios for each testing variation, totalling 144 iterations across all experiments and 45 hours of testing.

In terms of estimation and control, the RHMC module uses a horizon length of 3s, consisting of 15 sampling intervals of 0.2s, and the robot's steering angle is constrained to within $\pm 40^\circ$. The configuration of this module is otherwise the same as in \cite{Wallace_FSR}. The energy cost of motion model used for the global long-term planner is configured to use the `Bringelly' site configuration as defined in \cite{Wallace_WROCO}, which corresponds to the site from which the simulated operating environment was derived.

\section{RESULTS AND DISCUSSION}

Table \ref{tab:failures} summarises the performance of the robot in reaching the required positional accuracy for each waypoint. The \% of times that the robot is successfully able to navigate to the waypoint within timeouts of 1 minute and 5 minutes is shown for two different agent densities, 10$m^2$/agent and 50$m^2$/agent, simulating both a dense and sparse herd or crowd.  
It is clear that the choice of local planner impacts the robot's performance, with the failsafe only version failing to reach weeding waypoints (1m positional accuracy) in dense environments within the 1 min timeout 28\% of the time. This number drops significantly to 8.6\% in sparse environments, indicating the applicability of different planning methods to different environments. Versions of the framework using responsive local planners, MCTS and PF, perform well in both the sparse and dense environments, achieving only 2.8\% failure at 1m accuracy for MCTS and 4.6\% for PF for dense environments when restricted to 1 minute timeout, and less than 1\% failure when allowed to continue for up to 5 minutes. 

A key reason for these failures was due to the robot taking evasive action, which could occasionally perturb the robot pose to an extent that made subsequently reaching the goal region no longer kinematically feasible without taking a more circuitous route, which in turn increased the chance of further disruptions by the dynamic agents. This situation provides a clear motivation for the use of platforms with omnidirectional mobility---a direction we plan to pursue further with Swagbot once its sensing configuration supports it---as the increased mobility of such a platform would circumvent this situation almost entirely.

\begin{table}[]
\setlength\tabcolsep{4.8pt} 
\centering
\begin{tabular}{llllllll}
\hline
\multicolumn{1}{|l|}{\textbf{Timeout}} & \multicolumn{1}{c|}{\textbf{Local}}   & \multicolumn{3}{c|}{Dense (10 $m^2$/agent)}                                                                              & \multicolumn{3}{c|}{Sparse (50 $m^2$/agent)}                                                                              \\ \cline{3-8} 
\multicolumn{1}{|l|}{}                                  & \multicolumn{1}{l|}{\textbf{Planner}} & \multicolumn{1}{l|}{\textit{\textbf{5m}}} & \multicolumn{1}{l|}{\textit{\textbf{2m}}} & \multicolumn{1}{l|}{\textit{\textbf{1m }}} & \multicolumn{1}{l|}{\textit{\textbf{5m}}} & \multicolumn{1}{l|}{\textit{\textbf{2m}}} & \multicolumn{1}{l|}{\textit{\textbf{1m}}} \\ \hline \hline
\multicolumn{1}{|l|}{\textit{1min}}    & \multicolumn{1}{l|}{\textbf{MCTS}}    & \multicolumn{1}{l|}{0\%}           & \multicolumn{1}{l|}{0.4\%}         & \multicolumn{1}{l|}{2.8\%}         & \multicolumn{1}{l|}{0\%}           & \multicolumn{1}{l|}{0\%}           & \multicolumn{1}{l|}{0.4\%}         \\ \cline{2-8} 
\multicolumn{1}{|l|}{}                                  & \multicolumn{1}{l|}{\textbf{PF}}      & \multicolumn{1}{l|}{0\%}           & \multicolumn{1}{l|}{1.5\%}        & \multicolumn{1}{l|}{4.6\%}        & \multicolumn{1}{l|}{0\%}           & \multicolumn{1}{l|}{0.5\%}        & \multicolumn{1}{l|}{1.0\%}        \\ \cline{2-8} 
\multicolumn{1}{|l|}{}                                  & \multicolumn{1}{l|}{\textbf{FS}}      & \multicolumn{1}{l|}{2.4\%}        & \multicolumn{1}{l|}{8.5\%}        & \multicolumn{1}{l|}{28.0\%}       & \multicolumn{1}{l|}{0\%}           & \multicolumn{1}{l|}{3.7\%}        & \multicolumn{1}{l|}{8.6\%}        \\ \hline  \hline
\multicolumn{1}{|l|}{\textit{5mins}}   & \multicolumn{1}{l|}{\textbf{MCTS}}    & \multicolumn{1}{l|}{0\%}           & \multicolumn{1}{l|}{0.4\%}        & \multicolumn{1}{l|}{0.6\%}         & \multicolumn{1}{l|}{0\%}           & \multicolumn{1}{l|}{0\%}           & \multicolumn{1}{l|}{0\%}           \\ \cline{2-8} 
\multicolumn{1}{|l|}{}                                  & \multicolumn{1}{l|}{\textbf{PF}}      & \multicolumn{1}{l|}{0\%}           & \multicolumn{1}{l|}{0.5\%}        & \multicolumn{1}{l|}{0.8\%}        & \multicolumn{1}{l|}{0\%}           & \multicolumn{1}{l|}{0\%}           & \multicolumn{1}{l|}{0.7\%}        \\ \cline{2-8} 
\multicolumn{1}{|l|}{}                                  & \multicolumn{1}{l|}{\textbf{FS}}      & \multicolumn{1}{l|}{0\%}           & \multicolumn{1}{l|}{2.9\%}        & \multicolumn{1}{l|}{9.8\%}        & \multicolumn{1}{l|}{0\%}           & \multicolumn{1}{l|}{0\%}           & \multicolumn{1}{l|}{1.2\%}        \\ \hline
\end{tabular}
\caption{\textit{\% of failures not reaching required accuracy (5, 2m or 1m) within timeout (1min or 5min), for dense (10 $m^2$/agent) and sparse (50 $m^2$/agent) environments, with average distances of 3m and 7m between agents respectively.}\strut}
\label{tab:failures}
\end{table}

\begin{table}[]
\centering
\begin{tabular}{lllll}
\hline
\multicolumn{1}{|c|}{\textbf{Local}}   & \multicolumn{2}{c|}{Dense (10 $m^2$/agent)}                                                                    & \multicolumn{2}{c|}{Sparse (50 $m^2$/agent)}                                                                    \\ \cline{2-5} 
\multicolumn{1}{|l|}{\textbf{Planner}} & \multicolumn{1}{l|}{\textit{\textbf{med (m)}}} & \multicolumn{1}{l|}{\textit{\textbf{std (m)}}} & \multicolumn{1}{l|}{\textit{\textbf{med (m)}}} & \multicolumn{1}{l|}{\textit{\textbf{std (m)}}} \\ \hline \hline
\multicolumn{1}{|l|}{\textbf{MCTS}}    & \multicolumn{1}{l|}{1.63}                     & \multicolumn{1}{l|}{22.57}                 & \multicolumn{1}{l|}{0.28}                      & \multicolumn{1}{l|}{13.5}                   \\ \hline
\multicolumn{1}{|l|}{\textbf{PF}}      & \multicolumn{1}{l|}{3.9}                      & \multicolumn{1}{l|}{37.50}                  & \multicolumn{1}{l|}{0.20}                      & \multicolumn{1}{l|}{12.26}                   \\ \hline
\multicolumn{1}{|l|}{\textbf{FS}}      & \multicolumn{1}{l|}{0.05}                     & \multicolumn{1}{l|}{5.13}                   & \multicolumn{1}{l|}{0.06}                      & \multicolumn{1}{l|}{3.05}                   \\ \hline
\end{tabular}\caption{\textit{Displacement from optimal path, shown for all simulated tests, in both  dense and sparse densities of dynamic agents. Median and standard deviation are shown in metres for each tested local path planner.}\strut}
\label{tab:dist2path}
\end{table}

\begin{table}[]
\setlength\tabcolsep{4.5pt} 
\centering
\begin{tabular}{lllllll}
\hline
\multicolumn{1}{|c|}{\textbf{Local}}         & \multicolumn{3}{c|}{Dense (10 $m^2$/agent)}                                                                                      & \multicolumn{3}{c|}{Sparse (50 $m^2$/agent)}                                                                                      \\ \cline{2-7} 
\multicolumn{1}{|l|}{\textbf{Planner}}       & \multicolumn{1}{l|}{\textit{\textbf{\% coll}.}} & \multicolumn{1}{l|}{\textit{\textbf{med (m)}}} & \multicolumn{1}{l|}{\textit{\textbf{std (m)}}} & \multicolumn{1}{l|}{\textit{\textbf{\% coll}.}} & \multicolumn{1}{l|}{\textit{\textbf{med (m)}}} & \multicolumn{1}{l|}{\textit{\textbf{std (m)}}} \\ \hline \hline
\multicolumn{1}{|l|}{\textbf{MCTS}} & \multicolumn{1}{l|}{4.70}              & \multicolumn{1}{l|}{2.61}             & \multicolumn{1}{l|}{1.07}          & \multicolumn{1}{l|}{0.71}              & \multicolumn{1}{l|}{3.81}             & \multicolumn{1}{l|}{1.72}          \\ \hline
\multicolumn{1}{|l|}{\textbf{PF}}   & \multicolumn{1}{l|}{5.07}              & \multicolumn{1}{l|}{2.54}             & \multicolumn{1}{l|}{1.00}          & \multicolumn{1}{l|}{0.62}              & \multicolumn{1}{l|}{3.78}             & \multicolumn{1}{l|}{1.78}          \\ \hline
\multicolumn{1}{|l|}{\textbf{FS}}   & \multicolumn{1}{l|}{3.25}              & \multicolumn{1}{l|}{2.63}             & \multicolumn{1}{l|}{0.95}          & \multicolumn{1}{l|}{1.34}              & \multicolumn{1}{l|}{3.62}             & \multicolumn{1}{l|}{1.63}          \\ \hline
\end{tabular}
\caption{\textit{Minimum distance to agents, for all tests, in both dense and sparse environments. \% of collisions (dist \textless 2m) per interaction is shown with median and standard deviation in metres for each tested local path planner.}\strut}
\label{tab:dist2agents}
\end{table}

Fig. \ref{avgspeed_vs_posacc} illustrates the average speed at which the robot was able to reach the required positional accuracy of each planned waypoint for each test (within 5m for observation waypoints, 2m for soil sampling and 1m for weeding). The MCTS local planner implementation was able to outperform both the PF and FS versions for all accuracies, in both sparse and dense agent densities.  These results indicate that in the presence of dynamic agents, the time efficiency of the system is greatly dependent on the type of local planner used. The robot is able to complete the required mission objectives in significantly less time when a more responsive local planner is included in the framework. 
However, the results shown in Table \ref{tab:dist2path} show that the choice of local planner also impacts the robot's energy efficiency. Deviation from the offline optimal path is used as a proxy for energy use, indicating that a less responsive local planner---in this case the FS method used alone---uses significantly less energy than a planner which adapts its path to account for the motion of nearby agents.

Additionally, Table \ref{tab:dist2agents} shows that whilst all tested versions are usually able to maintain adequate distance from agents, with very similar median distances and standard deviations for both agent environment densities, the non-responsive FS method results in significantly more collisions with agents in the sparse environment. This result suggests that a number of collisions are due to the agents running into the robot, rather than the robot hitting the agents, and can be avoided by using a more responsive planner in sparse environments.
Conversely, in the dense environment, the more responsive versions result in greater number of collisions, suggesting that when the environment becomes too complex these methods may in fact cause more collisions through their efforts to avoid an initial collision. Our results suggest that by using various combinations of planners, the framework is able to achieve different performances in varied environments, demonstrating not just the flexibility of the framework to adapt to changes in the environment, such as density of crowds or herds, but also to changing requirements, such as energy usage, or speed of mission completion.




\section{CONCLUSION}
This work has proposed a hierarchical framework enabling the long-term operation of a field robot for navigation of mission waypoints in dynamic environments whilst adhering to resource budgets and other operational constraints. In particular, we have shown how this framework can handle changes in the environment and in current mission constraints, by using an adaptable local planning module to achieve varied performance as required. We have evaluated the strategy in extensive simulated trials and demonstrated its capability to robustly adapt long term mission plans in the presence of moving individuals and obstacles for real world applications. In future work, we will apply this framework to a robot to test its capability in the continuous completion of tasks such as weeding, soil sampling, and observing crops and livestock.

\end{document}